\documentclass[11pt]{article}
\usepackage[hyperref]{acl}
\usepackage{times}
\usepackage{latexsym}
\usepackage{microtype}
\usepackage{amsmath}
\usepackage{amssymb}
\usepackage{booktabs}
\usepackage{graphicx}
\usepackage{multirow}
\usepackage{xcolor}
\usepackage{placeins}   
\graphicspath{{figures/}}

\makeatletter
\@ifpackageloaded{lineno}{\setlength{\linenumbersep}{3mm}}{}
\makeatother

\title{Geometric Deviation as an Unsupervised Pre-Generation\\
Reliability Signal: Probing LLM Representations for Answerability}

\author{
  Yucheng Du \\
  University of Southern California \\
  \texttt{yuchengd@usc.edu}
}

\begin{document}
\maketitle

\begin{abstract}
A reliable language model should be able to signal, prior to generation,
when a query falls outside its knowledge.
We investigate whether representation geometry can provide such a
\emph{pre-generation} signal by measuring the deviation of hidden states
from an answerable reference set—requiring no labeled failure data
and no access to model outputs.

Across three instruction-tuned models (Llama~3.1-8B, Qwen~2.5-7B,
and Mistral-7B-Instruct) and three prompt forms
(\textsc{Math}, \textsc{Fact}, \textsc{Code}),
we find that geometry primarily encodes \emph{task form}.
Within mathematical prompts, unanswerable inputs consistently deviate
from the answerable centroid, yielding strong separation
(ROC-AUC $0.78$--$0.84$).
This single-pass pre-generation signal outperforms a simple refusal
baseline and compares favorably to self-consistency.
It also captures cases where models do not explicitly refuse.

In contrast, no reliable geometric signal emerges for factual prompts,
indicating that the effect is form-conditional rather than universal.
Code prompts show large effect sizes with higher variance, suggesting
partial generalization beyond mathematical form.

A layer-wise analysis reveals that the signal arises in early layers
and gradually attenuates toward the output.
These results suggest that answerability-related geometry is established
before the final stages of generation.
Together, these findings indicate that geometric deviation can serve as
a lightweight \emph{pre-generation} signal that is reliable in structured
domains with formal answerability constraints, with clear boundaries
on where it generalizes.
\end{abstract}

\section{Introduction}

Hallucination—the generation of confident but incorrect responses—remains
a central reliability challenge for deployed language models
\citep{ji2023survey}.
Detecting likely failures \emph{before} generation is particularly valuable:
a pre-generation signal can trigger abstention or human review without
adding latency to the decoding process.
Prior work has approached reliability estimation through uncertainty
calibration \citep{kadavath2022language},
internal probing classifiers \citep{slobodkin-etal-2023-curious},
representation steering \citep{li2023inferencetime},
and supervised internal-state analysis
\citep{zhang-etal-2025-prompt,zhang-etal-2025-mhad}.
However, these methods either rely on labeled training data,
require access to model outputs, or are sensitive to model-specific
characteristics.
Whether \emph{unsupervised} representation geometry alone can function as
a practical pre-generation reliability signal—without labeled failure
data or access to model outputs—remains underexplored.

We investigate a minimal approach: measuring each prompt's cosine distance
from the centroid of answerable-class representations, requiring no labeled
failure data, no fine-tuning, and no output sampling.
Our design isolates answerability from confounding surface variation
using \textbf{matched pairs}, where each unanswerable prompt shares
the domain, length, and syntactic form of a corresponding answerable prompt,
differing only in the property causing unanswerability.
We validate across \textbf{three architecturally distinct
instruction-tuned models} at the same scale (7--8B parameters):
Llama~3.1-8B-Instruct \citep{dubey2024llama},
Qwen~2.5-7B-Instruct \citep{qwen2025qwen25technicalreport},
and Mistral-7B-Instruct-v0.3 \citep{jiang2023mistral}.
Holding scale constant while varying architecture and alignment recipe
lets us distinguish input-driven geometric signals from model-specific
artefacts.

Our main findings are:
(1)~Within mathematical form, geometric deviation yields strong separation
(ROC-AUC $0.78$--$0.84$ across all three models), outperforming a simple
refusal baseline and a multi-sample self-consistency baseline requiring
$5{\times}$ the inference cost—including cases that refusal-based detection
may not capture;
(2)~The effect is \emph{form-conditional}: within factual form, no
significant signal emerges across any of the three models,
establishing a principled boundary;
(3)~Within code form, large effect sizes appear across all three models,
though statistical significance is mixed at the sample size studied,
suggesting the phenomenon may extend beyond mathematical form;
(4)~A layer-wise analysis reveals the signal \emph{peaks at early layers}
and generally decreases toward the output layer, consistent across all
three models, suggesting answerability-related geometry is established
early in the network;
(5)~Strong cross-model geometric consensus on a subset of \textsc{Math-U}
prompts suggests the signal reflects input structure rather than
model-specific geometry;
(6)~Behavioral responses to geometric outliers diverge across models—
Qwen refuses, Llama does not explicitly refuse on the same prompts—
indicating that alignment training shapes how models \emph{act on}
geometric information, not the information itself.

\section{Background}

\paragraph{Representation geometry in LLMs.}
LLM hidden states exhibit strong \emph{anisotropy}: representations
cluster near a dominant direction, inflating pairwise cosine similarity
even for unrelated inputs \citep{ethayarajh-2019-contextual}.
Mean-centering removes this dominant direction and restores discriminability
\citep{godey-etal-2024-anisotropy}.
Prior work shows that structural linguistic information is geometrically
encoded in Transformer representations \citep{hewitt-manning-2019-structural},
and that task-specific function vectors emerge in later layers of
instruction-tuned models \citep{todd2023function},
motivating the view that geometry can reflect semantic properties
beyond surface form.

\paragraph{Reliability signals in LLM representations.}
\citet{kadavath2022language} show that LLMs are well-calibrated on
multiple-choice tasks.
\citet{slobodkin-etal-2023-curious} probe for answerability in reading
comprehension via supervised classifiers on context-dependent questions.
\citet{burns2022discovering} extract truth directions via contrastive
activation differences; \citet{li2023inferencetime} show that steering
attention heads can elicit truthful answers.
A recent survey by \citet{xia-etal-2025-survey} organises uncertainty
estimation approaches across four paradigms.
The most closely related work is PRISM \citep{zhang-etal-2025-prompt}
and MHAD \citep{zhang-etal-2025-mhad}, which use supervised probing
on internal states for hallucination detection across multiple layers.
PRISM trains a prompt-guided classifier on labeled hallucination examples
to identify factual errors at inference time;
MHAD performs deep multi-layer representation analysis using supervised
training signals derived from factuality annotations.
Our work differs in three respects: it requires no labels on failure or
unanswerable instances (only a reference set of answerable prompts,
available by construction in structured query domains),
it operates strictly before generation (no output tokens needed),
and it uses matched-pair construction to explicitly disentangle surface
form from answerability—enabling a controlled characterisation of
\emph{when} and \emph{where} geometric reliability signals arise,
rather than learning to discriminate post-hoc from labeled failures.

\paragraph{Layer-wise signal in Transformers.}
Probing studies have found that different linguistic properties peak at
different layers: syntactic information tends to emerge in middle layers,
while semantic and task-level information concentrates in later layers
\citep{hewitt-manning-2019-structural}.
Our layer-wise analysis adds to this literature by showing that
answerability geometry—a reliability-relevant property—peaks
\emph{unusually early} (layers 2--5), suggesting that the network
encodes input-level structural violations before committing to a generation
strategy in deeper layers.

\paragraph{Hallucination and output-level baselines.}
Semantic entropy-based methods detect hallucinations from model outputs
without accessing internal states \citep{farquhar2024detecting}.
We compare against a lightweight output-level refusal baseline,
representing the information available from generation alone.
Instruction-tuned models differ in their tendency to refuse versus
hallucinate on unanswerable inputs \citep{bai2022training},
a distinction we investigate empirically across three models.

\section{Experimental Setup}
\label{sec:setup}

\paragraph{Models.}
We use three instruction-tuned models at the same scale (7--8B parameters):
Llama~3.1-8B-Instruct \citep{dubey2024llama},
Qwen~2.5-7B-Instruct \citep{qwen2025qwen25technicalreport},
and Mistral-7B-Instruct-v0.3 \citep{jiang2023mistral},
all loaded via HuggingFace Transformers \citep{wolf2020transformers} in
\texttt{float16} precision on Apple Silicon MPS.
Holding scale constant isolates architectural and alignment recipe differences.
Mistral's training recipe differs from both Llama and Qwen, providing a
third alignment data point.

\paragraph{Representation extraction.}
For each prompt, we extract last-layer hidden states, apply mean pooling
over all input tokens, and subtract the global mean vector computed over
all prompts in a given run \citep{godey-etal-2024-anisotropy}.
All distances are cosine distances ($1 - \cos\theta$).
For the layer-wise analysis (Section~\ref{sec:layerwise}), we extract
mean-pooled hidden states at every layer (including the embedding layer),
yielding a matrix of shape $(n_{\text{prompts}}, n_{\text{layers}}, d)$.

\paragraph{Prompt forms and matched-pair construction.}
We study three prompt forms.

\textsc{Math} ($n = 50$ pairs): well-defined arithmetic, algebra, or
combinatorics questions (\textsc{Math-A}) paired with structurally
identical variants in which a defined quantity is replaced by an
undefined one (\textsc{Math-U}).
Unanswerability sources include: mathematically undefined operations
(e.g., $\sqrt{-169}$ in the reals; $\log_1 10$; $0^0$),
extremal impossibilities (e.g., ``the largest prime'';
``the last Fibonacci number''; ``the product of all positive integers''),
and unknown-quantity substitutions
(e.g., ``the current number of active volcanoes'').
Each pair preserves domain, syntactic structure, and approximate length;
the sole change is the introduction of the undefined element.

\textsc{Fact} ($n = 10$ pairs): verifiable factual questions paired with
variants referencing unknowable future events, non-existent entities,
or counterfactual premises.
Examples: ``capital of France'' / ``capital of France in 2050'';
``currency of Japan'' / ``currency of Atlantis.''

\textsc{Code} ($n = 30$ pairs): Python expression questions with
deterministic return values (\textsc{Code-A}) paired with structurally
identical variants (\textsc{Code-U}) whose evaluation is undefined,
raises a well-typed exception, or requires unbounded computation.
Examples: \texttt{max([3,1,4])} / \texttt{max([])}
(well-defined / raises \texttt{ValueError});
\texttt{sum([1,2,3])} / \texttt{sum(itertools.count())}
(finite / non-terminating);
\texttt{hash(42)} / \texttt{hash([1,2,3])}
(hashable / \texttt{TypeError}).
The \textsc{Code} form tests whether the geometric signal generalizes
beyond the mathematical domain to a domain where unanswerability arises
from type violations and semantic ill-definedness in a programming language.

In all three forms, construction rules are applied consistently:
one element is changed per pair; surface structure is preserved.
This design rules out length, domain, and surface form as confounds.
All prompts and analysis code are released at
\url{https://github.com/yucheng-du/geom-reliability}.

\paragraph{Analysis.}
For all controlled experiments, we compute each prompt's \emph{own\_dist}—
cosine distance to its form's \textsc{A}-only centroid—as the reliability score.
Centroids are computed from the \textsc{A}-labeled prompts only,
so no \textsc{U}-label information enters the score construction.
In a deployment setting this reference set corresponds to a small collection
of prompts known to be answerable (e.g., standard queries in a domain),
requiring no annotation of failures or unanswerable instances.
We report one-sided permutation tests ($n_{\text{perm}} = 5000$) on the
mean gap $\overline{\text{dist}}_U - \overline{\text{dist}}_A$,
recomputing the centroid at each permutation to avoid null-hypothesis
violations, together with Cohen's $d$ for effect size.
Mean-centering is performed jointly over all prompts within a run.
For the \textsc{Math}/\textsc{Fact} experiments, \textsc{Fact} and
\textsc{Math} prompts are mean-centered together and share the same
representational reference frame.
The \textsc{Code} experiments were run separately and use their own
mean-centering context; own\_dist values for \textsc{Code} are therefore
not directly comparable on an absolute scale to \textsc{Math}/\textsc{Fact}
values.

For reliability prediction, we threshold own\_dist at the midpoint of the
mean \textsc{A} and mean \textsc{U} distances to produce a binary classifier
and report ROC-AUC and F1.
The refusal-keyword baseline classifies a prompt as unanswerable if the
model's generated output contains any of a curated list of
refusal-indicative surface tokens:
\textit{undefined}, \textit{cannot}, \textit{doesn't exist},
\textit{no such}, \textit{not defined}, \textit{infinite},
\textit{ValueError}, \textit{TypeError}, \textit{ZeroDivisionError},
and related forms.
This baseline represents the information extractable from the model's
output alone—requiring a completed generation pass—and serves as a
practical upper bound for lightweight output-level detection.
Its recall is structurally bounded: it can only fire when the model
explicitly names its uncertainty, and cannot detect hallucinations
where the model generates confidently incorrect responses without
refusal markers.

We additionally evaluate a \textbf{self-consistency} (SC) baseline:
for each prompt, we generate $k=5$ samples at temperature $0.7$ and
compute a disagreement score.
For \textsc{Math} and \textsc{Code}, we extract the final answer token
from each sample and set the score to $1 - (\text{majority count}/k)$
(\textit{answer\_disagree}); for \textsc{Fact}, where answers are
free-form, we compute the mean pairwise ROUGE-1 F1 over the last-line
excerpts of all $\binom{k}{2}$ sample pairs and set the score to
$1 - \overline{\text{ROUGE-1}}$ (\textit{rouge\_disagree}).
SC requires five generation passes per prompt and accesses model outputs.
We note that this disagreement-based SC is a lightweight proxy, not full
semantic entropy \citep{farquhar2024detecting}: it relies on surface string
matching of extracted answer tokens rather than semantic clustering across
outputs, and thus constitutes a lower bound on what output-level uncertainty
estimation can achieve; full semantic entropy remains future work.
We include SC to characterise how a post-generation multi-sample baseline
compares to the single-pass pre-generation geometry signal.

\section{Results}

\subsection{Geometry Encodes Task Form}
\label{sec:taskform}

Llama and Qwen produce well-separated clusters for the three prompt categories
in the uncontrolled task-structure experiment (all $p < 0.01$,
permutation test on within- vs.\ between-group cosine distances).
\textsc{Math} forms the tightest cluster
(within-class distance: Llama $0.332$, Qwen $0.415$),
reflecting the high surface uniformity of arithmetic questions.
Centroid analysis reveals an asymmetry: the \textsc{Fact}--\textsc{Math}
centroid cosine ($\approx -0.84$ to $-0.85$) indicates near-orthogonality
after mean-centering, while \textsc{Fact}--\textsc{Unknown} is positive
and moderately close ($+0.41$ Llama, $+0.58$ Qwen).
The \textsc{Unknown} cluster therefore aligns with \textsc{Fact}, not \textsc{Math}—
a pure form effect: math-form unanswerable prompts are pulled toward
the \textsc{Math} centroid, while fact-form unanswerable prompts align
with \textsc{Fact}.

\begin{figure*}[!t]
  \centering
  \includegraphics[width=\textwidth]{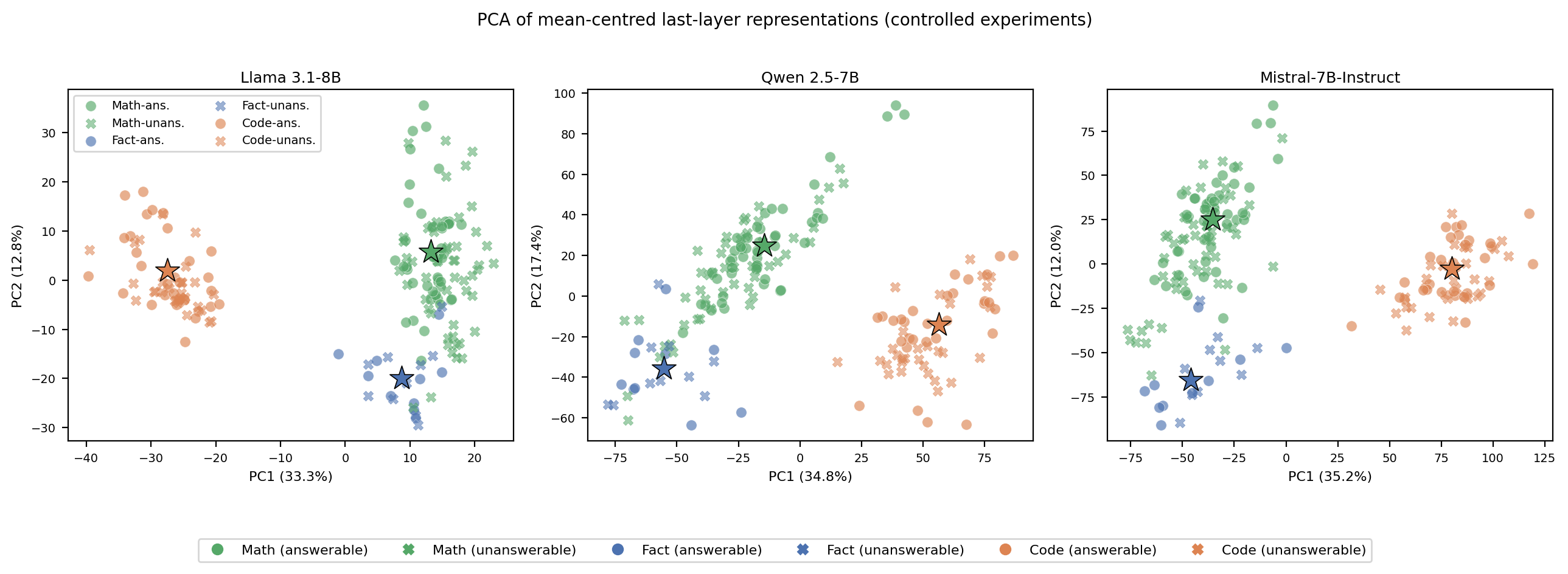}
  \caption{PCA of mean-centred last-layer representations (controlled experiments)
  for all three models.
  \textsc{Math} (green), \textsc{Fact} (blue), and \textsc{Code} (orange)
  occupy distinct geometric regions.
  Filled circles: answerable; crosses: unanswerable.
  Stars mark answerable-class centroids.
  \textsc{Math} forms the tightest cluster; \textsc{Code} is geometrically
  distant from \textsc{Math} and shows higher within-class spread,
  consistent with greater surface heterogeneity in Python expressions.
  The three-form separation is consistent across all three architectures.}
  \label{fig:pca}
\end{figure*}

In the controlled experiments, the \textsc{Code} form occupies a distinct
cluster well-separated from both \textsc{Math} and \textsc{Fact},
suggesting that programming language structure is encoded geometrically
in instruction-tuned representations distinctly from natural-language forms.
\textsc{Code} within-class distances are highest of the three forms
(Llama $\approx 0.889$, Qwen $\approx 0.815$, Mistral $\approx 0.875$),
reflecting greater surface heterogeneity in Python expressions
relative to arithmetic questions.
The \textsc{Code}--\textsc{Math} centroid distance is large (both forms
produce tight but geometrically distant clusters), whereas
\textsc{Code} and \textsc{Fact} exhibit intermediate separation.
This structure is consistent across all three models, as shown in
Figure~\ref{fig:pca}, suggesting that the task-form encoding is not
an artifact of a specific architecture.

\subsection{Answerability Signal Within Form}
\label{sec:answerability}

Table~\ref{tab:answerability} reports the full controlled answerability results
for all three models and all three forms.

\begin{table}[htbp]
\small
\centering
\resizebox{\columnwidth}{!}{%
\begin{tabular}{llrrrrc}
\toprule
\textbf{Form} & \textbf{Model} &
$n$ &
\textbf{dist\textsubscript{A}} &
\textbf{dist\textsubscript{U}} &
\textbf{$\Delta$} & \textbf{$p$} \\
\midrule
\multirow{3}{*}{\textsc{Math}}
  & Llama   & 50 & 0.676 & 1.055 & +0.379 & \textbf{$<$.0001} \\
  & Qwen    & 50 & 0.652 & 1.042 & +0.390 & \textbf{$<$.0001} \\
  & Mistral & 50 & 0.668 & 1.038 & +0.370 & \textbf{$<$.0001} \\
\midrule
\multirow{3}{*}{\textsc{Fact}}
  & Llama   & 10 & 0.326 & 0.406 & +0.080 & 0.498 \\
  & Qwen    & 10 & 0.303 & 0.361 & +0.058 & 0.566 \\
  & Mistral & 10 & 0.305 & 0.402 & +0.097 & 0.360 \\
\midrule
\multirow{3}{*}{\textsc{Code}}
  & Llama   & 30 & 0.889 & 1.195 & +0.306 & 0.113 \\
  & Qwen    & 30 & 0.815 & 1.229 & +0.414 & \textbf{0.008} \\
  & Mistral & 30 & 0.875 & 1.168 & +0.294 & 0.155 \\
\bottomrule
\end{tabular}}
\caption{Cosine distance to answerable centroid (own\_dist) for matched pairs.
$\Delta = \text{dist}_U - \text{dist}_A$;
$p$-values from one-sided permutation test ($n_\text{perm}=5000$).
\textsc{Math}/\textsc{Fact}: joint mean-centering;
\textsc{Code}: separate mean-centering context (see \S\ref{sec:setup}).}
\label{tab:answerability}
\end{table}

\begin{figure*}[!t]
  \centering
  \includegraphics[width=\textwidth]{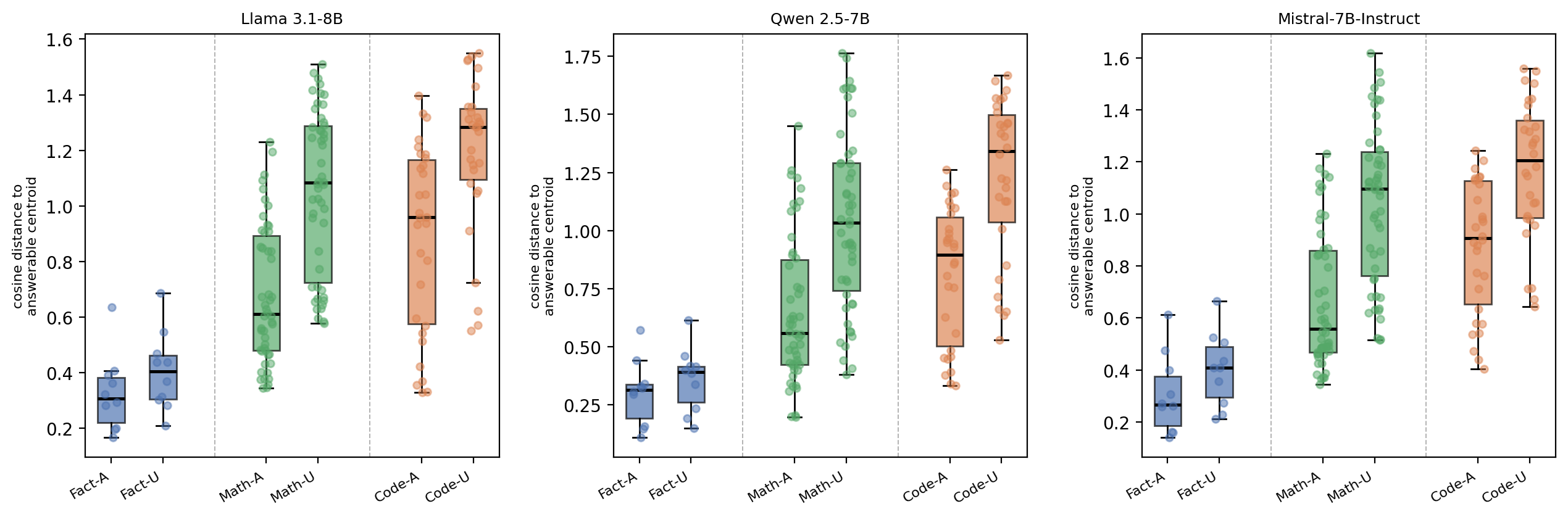}
  \caption{Distribution of own\_dist (cosine distance to answerable centroid)
  for all three models and all three prompt forms.
  \textsc{Math-U} and \textsc{Code-U} distributions are substantially elevated
  relative to their answerable counterparts, consistent across all three models.
  \textsc{Fact-A} and \textsc{Fact-U} distributions largely overlap,
  consistent with the non-significant permutation tests.
  \textsc{Code} distances are not directly comparable to \textsc{Math}/\textsc{Fact}
  values as they are computed under separate mean-centering contexts.
  Individual prompts shown as jittered points.}
  \label{fig:boxplot}
\end{figure*}

\paragraph{\textsc{Math} ($n = 50$ pairs).}
All three models show highly significant separation between \textsc{Math-A}
and \textsc{Math-U} at the expanded sample size ($p < 0.0001$,
Cohen's $d$ ranging from $+1.12$ to $+1.41$).
The \textsc{Math-U} centroid distance ($\approx 1.04$--$1.06$) is
substantially higher than \textsc{Math-A} ($\approx 0.65$--$0.68$),
a gap of $\approx +0.37$--$+0.39$ that is consistent across architectures.
We attribute this separation to the structural-contradiction hypothesis:
mathematical unanswerability forces a representation toward an undefined
region of the tight \textsc{Math} attractor, producing systematic
centroid deviation.
Figure~\ref{fig:boxplot} shows the full own\_dist distributions:
\textsc{Math-U} exhibits substantially elevated values with heavy tails,
while \textsc{Fact-A} and \textsc{Fact-U} distributions overlap
completely—directly visualising the null \textsc{Fact} result.

\paragraph{\textsc{Fact} ($n = 10$ pairs).}
No significant separation emerges for any model ($p = 0.36$--$0.57$;
Cohen's $d = +0.44$--$+0.76$).
The null result is not underpowered: at $n = 10$ pairs, the \textsc{Math}
effect was already $p < 0.01$ in the original experiments.
Factual unanswerability (future events, non-existent entities)
is syntactically indistinguishable from ordinary factual questions
and does not disrupt the \textsc{Fact} cluster geometry.

\paragraph{\textsc{Code} ($n = 30$ pairs).}
Effect sizes are large and consistent across models
($d = +1.01$, $+1.31$, $+1.14$ for Llama, Qwen, Mistral),
but statistical significance is mixed:
Qwen reaches $p = 0.008$; Llama and Mistral are $p = 0.11$--$0.16$.
The large $d$ values alongside marginal $p$-values indicate
higher within-group variance in the \textsc{Code} domain:
some \textsc{Code-A} prompts already occupy high-deviation positions
(e.g., those with unusual expression structure), widening the baseline
variance and reducing power relative to \textsc{Math}.
We interpret this as evidence that a similar phenomenon exists in the
\textsc{Code} domain but requires larger $n$ to reach conventional significance.

\subsection{Layer-wise Signal Profile}
\label{sec:layerwise}

\begin{figure*}[!t]
  \centering
  \includegraphics[width=0.92\textwidth]{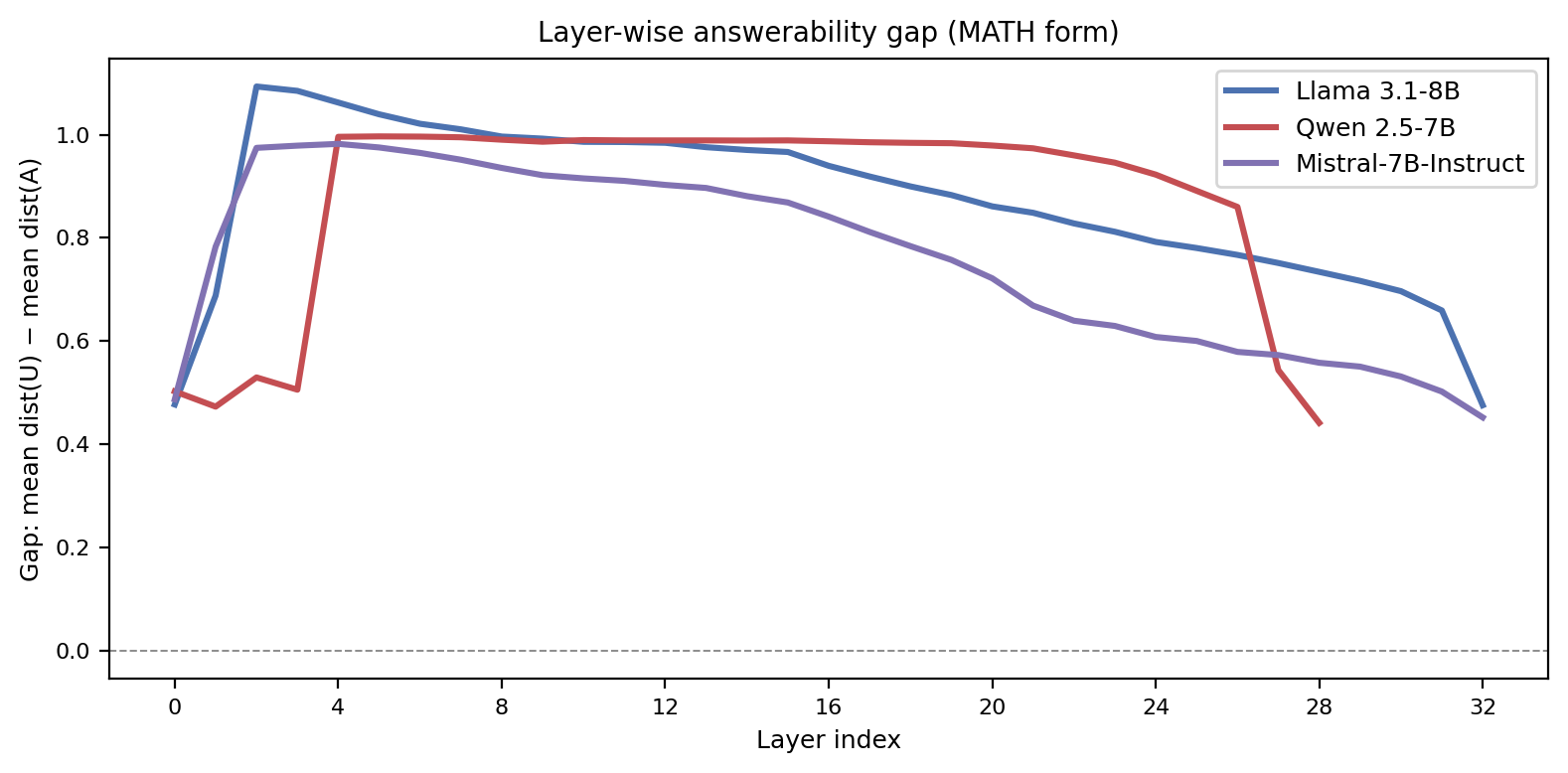}
  \vspace{4pt}
  \includegraphics[width=\textwidth]{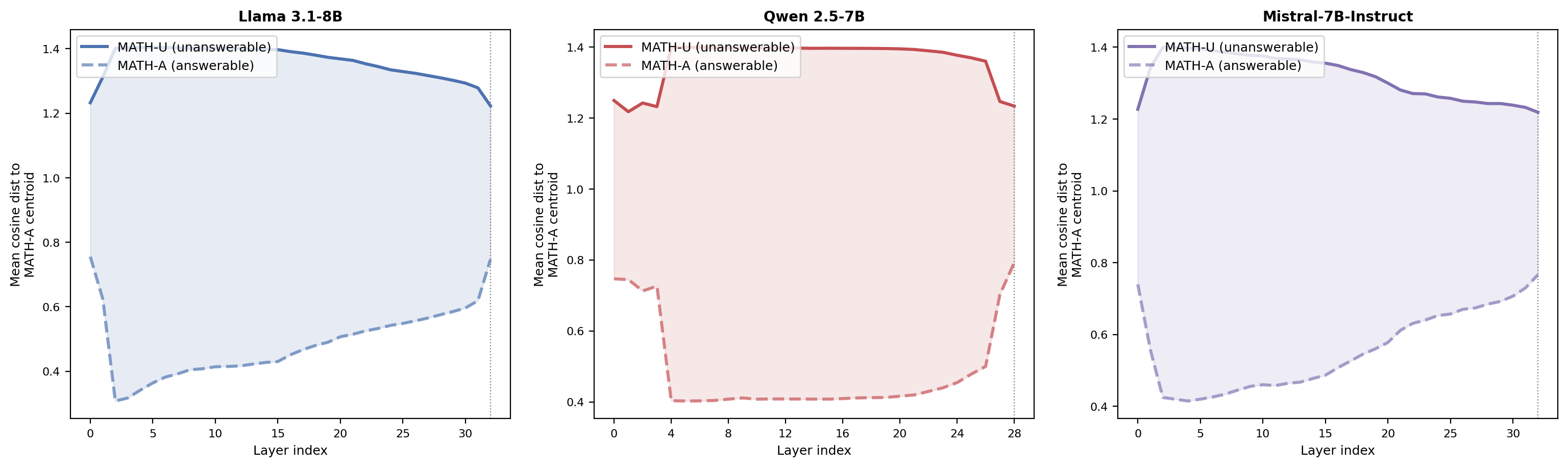}
  \caption{\textit{Top}: Layer-wise answerability gap $\delta_l$ for
  \textsc{Math} matched pairs ($n=20$). All three models peak at layers 2--5
  and generally decrease toward the last layer.
  \textit{Bottom}: Absolute own\_dist traces for \textsc{Math-U} (solid) and
  \textsc{Math-A} (dashed). The gap narrows because \textsc{Math-A} rises
  in deeper layers, not because the \textsc{Math-U} signal decays.}
  \label{fig:layerwise}
\end{figure*}

To understand \emph{where} in the Transformer stack the answerability signal
arises, we extract mean-pooled hidden states at every layer
for all 20 \textsc{Math} matched pairs and compute the
per-layer gap $\delta_l = \overline{\text{dist}}_U^{(l)} - \overline{\text{dist}}_A^{(l)}$
for all three models.

Figure~\ref{fig:layerwise} shows $\delta_l$ as a function of layer index.
The pattern is consistent across all three models:
the gap rises sharply from the embedding layer,
\textbf{peaks at an early layer} (layer 2 for Llama, layer 5 for Qwen,
layer 4 for Mistral; peak $\delta \approx 0.98$--$1.09$),
and then \textbf{generally decreases} through subsequent layers
to the final layer (last-layer $\delta \approx 0.44$--$0.48$).
The last layer retains a large, practically useful gap,
but is the \emph{minimum} among all middle layers—not the maximum.

Figure~\ref{fig:layerwise} (bottom) clarifies the mechanism underlying the gap profile:
\textsc{Math-U} representations diverge from the answerable centroid from
the earliest attention layers and sustain that distance throughout the
network.
The overall decrease in $\delta_l$ is driven by the answerable class
drifting \emph{toward} the unanswerable class as depth increases,
not by the unanswerable signal weakening.
This suggests the network progressively adapts \textsc{Math-A} representations
toward a generation-ready state that incidentally reduces their distance
to the answerable centroid, while \textsc{Math-U} representations
remain anchored in a structurally anomalous region.

This pattern indicates that answerability geometry is an emergent property
of the earliest attention layers, attenuating as the network adapts
representations toward generation readiness.
Using the last layer in our main experiments follows prior work convention;
the layer-wise profile suggests earlier layers could yield stronger classifiers
if layer selection were optimized (Section~\ref{sec:discussion}).

\subsection{Geometry--Behavior Alignment}
\label{sec:behavior}

We annotate model outputs for the original 20 \textsc{Math-U} prompts
(Table~\ref{tab:behavior}).

\begin{table}[htbp]
\small
\centering
\begin{tabular}{lrrr}
\toprule
\textbf{Model} & \textsc{Refuse} & \textsc{Partial} & \textsc{Halluc} \\
\midrule
Llama 3.1-8B & 5/20 & 5/20 & 10/20 \\
Qwen 2.5-7B  & 7/20 & 3/20 & 10/20 \\
\bottomrule
\end{tabular}
\caption{Behavioral annotation for \textsc{Math-U} prompts ($n=20$).
Annotation based on qualitative reading of generated output.}
\label{tab:behavior}
\end{table}
Both Llama and Qwen hallucinate on 10/20 prompts.
Qwen refuses more often (7/20 vs.\ 5/20 for Llama);
Llama partially answers more (5/20 vs.\ 3/20).

Geometric deviation predicts behavior within each model:
all four Llama prompts with own\_dist $> 1.2$ (drifted to the
\textsc{Fact} centroid) produce hallucination or partial answers,
with zero refusals.
For Qwen, the five \textsc{Refuse} cases cluster within the \textsc{Math}
cluster (non-drifted), while the two highest-deviation prompts
(m07u: ``next prime after the largest prime''; m10u: ``$\pi$ for a square'')
are correctly identified as undefined and refused.
The \textbf{critical divergence}: these same prompts show the highest
own\_dist in \emph{both} models, yet Llama hallucinates while Qwen refuses.

At the expanded scale ($n = 50$ \textsc{Math-U} prompts),
\textbf{19 of 50 prompts are misassigned to the \textsc{Fact} centroid
in all three models simultaneously}.
These 19 prompts share a common property: they involve extremal or
infinite mathematical objects (``the largest prime,'' ``the last Fibonacci number,''
``the average of all positive reals,'' ``the product of all positive integers'')
or operations on unknown future/unbounded quantities.
The near-perfect cross-model geometric consensus on these prompts—
across three different architectures and training recipes—provides strong evidence
that the signal reflects input-level structural properties rather than
any model-specific geometry.

The 19 consensus-drift prompts fall into three structural categories:
(i)~extremal or infinite objects (``the largest prime,'' ``the last Fibonacci number''),
(ii)~unbounded aggregates (``the exact sum of all natural numbers to infinity''),
and (iii)~unknown-quantity substitutions (``17 multiplied by the current moons of Jupiter'').
Categories (i) and (ii) account for most cases, consistent with representations
detecting \emph{formal} impossibility rather than epistemic difficulty.

The same-geometry, different-behavior pattern between Llama and Qwen on
shared outliers further supports the view that alignment training modulates
how models \emph{act on} geometric information, not the information itself:
the representation-level signal of ``this is anomalous'' is present
in all three models; the behavioral choice of whether to refuse, partially
answer, or hallucinate is model-specific.

\subsection{Reliability Prediction Evaluation}
\label{sec:auc}

Table~\ref{tab:auc} evaluates own\_dist as an unsupervised binary classifier
distinguishing answerable from unanswerable matched pairs,
compared against the refusal-keyword baseline.

\begin{table}[htbp]
\footnotesize
\centering
\resizebox{\columnwidth}{!}{%
\begin{tabular}{llrrrrr}
\toprule
\textbf{Form} & \textbf{Model} &
\multicolumn{2}{c}{\textbf{Geometry}} &
\multicolumn{1}{c}{\textbf{SC}} &
\multicolumn{2}{c}{\textbf{Refusal}} \\
\cmidrule(lr){3-4}\cmidrule(lr){5-5}\cmidrule(lr){6-7}
& & \textbf{AUC} & \textbf{F1} & \textbf{AUC} & \textbf{AUC} & \textbf{F1} \\
\midrule
\multirow{3}{*}{\textsc{Math}}
  & Llama   & \textbf{0.841} & \textbf{0.714} & 0.624 & 0.630 & 0.413 \\
  & Qwen    & \textbf{0.782} & \textbf{0.694} & 0.296 & 0.710 & 0.592 \\
  & Mistral & \textbf{0.826} & \textbf{0.714} & 0.524 & 0.730 & 0.630 \\
\midrule
\multirow{3}{*}{\textsc{Fact}}
  & Llama   & 0.690 & 0.632 & 0.460 & 0.550 & 0.182 \\
  & Qwen    & 0.660 & 0.700 & 0.000 & \textbf{0.750} & \textbf{0.667} \\
  & Mistral & 0.710 & 0.700 & 0.470 & 0.550 & 0.308 \\
\midrule
\multirow{3}{*}{\textsc{Code}}
  & Llama   & \textbf{0.774} & \textbf{0.758} & 0.441 & 0.633 & 0.421 \\
  & Qwen    & 0.818 & 0.733 & 0.369 & \textbf{0.850} & \textbf{0.830} \\
  & Mistral & \textbf{0.796} & \textbf{0.689} & 0.497 & 0.733 & 0.636 \\
\bottomrule
\end{tabular}}
\caption{ROC-AUC and F1 for answerability prediction.
\textbf{Geometry}: own\_dist (pre-generation, zero samples);
\textbf{SC}: self-consistency disagreement score (post-generation, 5 samples);
\textbf{Refusal}: keyword classifier on single generated output (post-generation).
\textsc{Math}: $n=50$ pairs; \textsc{Fact}: $n=10$ pairs;
\textsc{Code}: $n=30$ pairs.
F1 threshold: midpoint of mean A and mean U own\_dist (Geometry);
SC F1 omitted—oracle-threshold F1 is $0.667$ across all conditions,
indicating SC is a near-constant classifier on this task.}
\label{tab:auc}
\end{table}

\paragraph{\textsc{Math}.}
Geometry substantially outperforms both baselines across all three models.
Against the refusal baseline (AUC $0.63$--$0.73$), geometry achieves AUC
$0.78$--$0.84$: refusal suffers from low recall, as only a fraction of
\textsc{Math-U} prompts trigger explicit refusal keywords.
Against the disagreement-based SC baseline (AUC $0.30$--$0.62$),
the margin is even larger: instruction-tuned models tend to hallucinate
\emph{consistently}, producing the same incorrect answer across all five
samples and yielding near-zero disagreement despite high geometric deviation.
Geometry captures this pattern pre-generation, whereas SC—which detects
output variance—cannot distinguish confident hallucination from correct answers
in this regime; a stronger semantic entropy baseline could narrow this gap.

\paragraph{\textsc{Fact}.}
Geometry yields modest AUC ($0.66$--$0.71$), consistent with the
non-significant permutation tests.
Qwen's refusal baseline ($0.75$) outperforms geometry here,
reflecting Qwen's tendency to explicitly refuse future-event questions.
Llama and Mistral refusal baselines are near-chance ($0.55$),
reflecting their tendency to answer rather than refuse.
SC is weakest on \textsc{Fact}: Qwen SC AUC $= 0.000$, because Qwen
gives consistent refusal-style responses even to answerable factual
questions, eliminating any disagreement signal.
At $n = 10$ pairs, \textsc{Fact} AUC estimates carry high variance and
should be treated as exploratory; the permutation tests ($p > 0.34$
across all three models) provide the more reliable evidence that no
systematic geometric signal exists for factual unanswerability at
the sample sizes studied.

\paragraph{\textsc{Code}.}
For Llama and Mistral, geometry ($0.77$--$0.80$) outperforms both SC
($0.44$--$0.50$) and refusal ($0.63$--$0.73$).
Qwen is the exception: its refusal baseline ($0.85$) exceeds geometry ($0.82$),
because Qwen explicitly names exception types
(``this raises a \texttt{TypeError}'') in its outputs for ill-typed
\textsc{Code-U} prompts, making keyword detection highly informative.
This Qwen-specific behavior mirrors its elevated refusal rate in \textsc{Math}
and is consistent with Qwen's alignment recipe producing more explicit
uncertainty acknowledgment.
SC remains the weakest signal across all three models on \textsc{Code}
(AUC $0.37$--$0.50$), confirming that output variance alone cannot
reliably distinguish well-typed from ill-typed expressions when the model
generates plausible-sounding but incorrect responses.

\FloatBarrier
\section{Discussion}
\label{sec:discussion}

\paragraph{Form dominates; answerability disrupts only when structural.}
Task form appears to be the primary organiser of last-layer geometry.
Answerability appears as a secondary, \emph{conditional} signal:
it is most visible when unanswerability creates a structural inconsistency
within the form—applying arithmetic to an undefined quantity may push a
representation toward an unfamiliar region within the tight \textsc{Math}
cluster, whereas factual unanswerability (future events, counterfactuals)
is syntactically indistinguishable from ordinary factual questions and
leaves the geometric cluster intact.
The \textsc{Code} results are consistent with this view: ill-defined Python
expressions (type errors, non-terminating operations) show a large geometric
effect, though more data are needed to confirm it reaches conventional
significance.
The \textsc{Math}/\textsc{Fact} asymmetry is consistent with this
form-attractor account.

\paragraph{Answerability signal is an early-layer phenomenon.}
The layer-wise profile—peaking at layers 2--5 and generally decreasing
thereafter—contrasts with prior probing work that finds semantic properties strongest
in later layers \citep{hewitt-manning-2019-structural}.
The attenuation is asymmetric (Figure~\ref{fig:layerwise}, bottom):
\textsc{Math-U} distances remain elevated throughout; \textsc{Math-A}
distances rise in deeper layers, narrowing the gap from below rather than above.
Deeper layers do not erase the anomaly signal but ``normalise'' answerable
inputs toward a generation manifold, trading geometric separability for
generation readiness.
For early-warning systems, layer-2--5 activations may yield stronger
answerability classifiers than the final layer at the cost of monitoring
intermediate activations.

\paragraph{Geometry as a form-conditional reliability indicator.}
The \textsc{Math} AUC results ($0.78$--$0.84$ across three models) establish
that unsupervised geometric deviation is a viable pre-generation reliability
signal in settings where unanswerability disrupts form structure.
The refusal baseline can only detect failure \emph{after} the model has
decided to refuse; geometry operates over the full distribution of
unanswerable inputs, including those the model does not explicitly refuse.
Cross-model geometric consensus on a subset of \textsc{Math-U} prompts
further suggests that the signal reflects input-level structure rather
than model-specific geometry, widening potential applicability.
The self-consistency comparison sharpens this point: even with $5{\times}$ the
inference cost and post-generation access, SC achieves AUC of only
$0.30$--$0.62$ on \textsc{Math}—substantially below geometry ($0.78$--$0.84$).
The failure mode is systematic: instruction-tuned models hallucinate
\emph{consistently}, producing the same wrong answer across all five samples
and thus yielding near-zero disagreement.
Geometry captures structural anomaly \emph{before generation} on these
consistently-answered prompts, where output variance is uninformative.
Outside structurally disruptive settings, output-level signals such as
semantic entropy \citep{farquhar2024detecting} or calibrated confidence
\citep{kadavath2022language} may be more informative.

\paragraph{Alignment shapes the geometry--behavior link.}
The Llama/Qwen divergence on geometrically anomalous prompts—identical
geometry, divergent behavior—suggests that alignment training shapes
how models \emph{respond to} geometric anomaly signals rather than
altering the signals themselves \citep{bai2022training}.
All three models agree geometrically on the 19-prompt consensus set
even while differing in behavioral response.
Directly testing this base-versus-instruction-tuned hypothesis remains
future work.

\section{Conclusion}

We show that geometric deviation from an answerable reference set can
serve as a pre-generation signal for answerability, particularly when
unanswerability introduces structural inconsistencies within a prompt's form.
Across three instruction-tuned models, the signal yields strong separation
on \textsc{Math} (ROC-AUC $0.78$--$0.84$), no reliable signal on \textsc{Fact},
and large but variable effects on \textsc{Code}---a form-conditional pattern.
A layer-wise analysis localises the signal to early layers (2--5), suggesting
answerability-related geometry is established before the final stages of
generation, with strong cross-model consensus on which prompts are anomalous.

The approach requires no labeled failure data, operates prior to generation,
and is consistent across architecturally distinct models.
The form-dependence---strongest for structurally disruptive unanswerability,
weaker for open-domain factual queries---motivates combining geometric and
output-based reliability signals in future work.


\section{Limitations}

\textbf{Scale.}
Sample sizes are modest by benchmark standards
($n = 50$ matched pairs for \textsc{Math},
$n = 10$ for \textsc{Fact}, $n = 30$ for \textsc{Code}).
\textsc{Fact} and \textsc{Code} AUC estimates carry high variance.
The \textsc{Code} results in particular—large effect sizes but mixed
significance—suggest the phenomenon exists but requires larger $n$
(we estimate $n \gtrsim 80$--$100$ pairs) to reach conventional significance.
Future work should validate on larger matched-pair sets and held-out benchmarks
with independently verified answerability labels.

\textbf{Baselines.}
We compare against a refusal-keyword proxy and a self-consistency disagreement
baseline (5 samples per prompt).
SC is substantially weaker than geometry on \textsc{Math}
(AUC $0.30$--$0.62$ vs.\ $0.78$--$0.84$), confirming that output variance
is insufficient for this failure mode.
However, stronger baselines remain untested: full semantic entropy
\citep{farquhar2024detecting}, which clusters semantically equivalent
outputs rather than surface-identical answers, and supervised probing
methods such as PRISM \citep{zhang-etal-2025-prompt} and
MHAD \citep{zhang-etal-2025-mhad} require labeled training data but
may outperform our unsupervised signal on some form-condition combinations.

\textbf{Layer selection.}
Our main results use the last layer for comparability with prior work.
The layer-wise analysis (Section~\ref{sec:layerwise}) shows that
earlier layers (2--5) carry a stronger signal.
Systematic comparison of pooling strategies (last layer, early layer,
CLS token, last-token) across tasks is needed to identify optimal
representation extraction.

\textbf{Annotation.}
Behavioral outputs (Section~\ref{sec:behavior}) are labeled by a single
annotator; inter-annotator agreement was not measured, and Mistral
behavioral annotation was not performed.
A follow-up study with multiple annotators would strengthen the
geometry--behavior analysis.

\textbf{Alternative explanations.}
Two confounds cannot be fully excluded: (i)~\textsc{Math-U} and
\textsc{Code-U} prompts may be lexically unusual independently of
answerability, producing out-of-distribution representations;
(ii)~higher own\_dist may partly reflect greater intra-class variance
rather than a systematic centroid shift.
Controlling for perplexity is a concrete next step.

\textbf{Model and alignment scope.}
All three models are in the 7--8B range; scaling behavior and the
impact of different alignment recipes (comparing base and RLHF-tuned
variants of the same model family) remain open questions.

\textbf{Probe choice.}
Our negative result on \textsc{Fact} relies on a single unsupervised probe:
cosine distance to the answerable-class centroid on last-layer mean-pooled
representations.
We do not evaluate whether more sophisticated probes---e.g.\
PCA-projected directions, learned hyperplanes, or activations pooled from
earlier layers---recover a signal on factual unanswerability at the sample
size studied.
The form-conditional pattern we report may therefore partly reflect this
probe choice rather than an intrinsic property of the representations.
Characterising how probe complexity trades off against form-domain coverage
is left to future work.

\section{Broader Impact}

This work investigates whether internal representation geometry of LLMs
can serve as an unsupervised reliability signal, with potential applications
in hallucination detection and deployment safety monitoring.
A pre-generation signal that fires before output is produced could
complement generation-based uncertainty methods, particularly in
latency-sensitive or safety-critical settings.

Our findings are encouraging but bounded: the geometric signal is
reliable for structurally disruptive answerability failures
(mathematical undefined operations, ill-typed code expressions)
but not for general factual unanswerability.
Practitioners should not deploy geometric deviation as a universal
hallucination detector based on these results alone.

The layer-wise finding—that early layers carry stronger answerability
signals than the last layer—suggests that lightweight online monitoring
of intermediate activations could serve as a more efficient pre-generation
filter than full forward-pass representation extraction.

We note that representation probing methods, including ours, could
in principle be used adversarially—for example, to construct prompts
that bypass geometric detection while still causing hallucinations.
However, the specificity of the signal to structural form disruption
limits this concern in practice.
No personal data was used in this study; all prompts are researcher-constructed.

\bibliography{references}

@inproceedings{wolf2020transformers,
  title     = {Transformers: State-of-the-Art Natural Language Processing},
  author    = {Wolf, Thomas and Debut, Lysandre and Sanh, Victor and Chaumond, Julien and Delangue, Clement and Moi, Anthony and Cistac, Pierric and Rault, Tim and Louf, R{\'e}mi and Funtowicz, Morgan and Brew, Jamie},
  booktitle = {Proceedings of the 2020 Conference on Empirical Methods in Natural Language Processing: System Demonstrations},
  year      = {2020},
  pages     = {38--45},
  publisher = {Association for Computational Linguistics},
}

@inproceedings{ethayarajh-2019-contextual,
  title     = {How Contextual are Contextualized Word Representations? Comparing the Geometry of {BERT}, {ELMo}, and {GPT}-2 Embeddings},
  author    = {Ethayarajh, Kawin},
  booktitle = {Proceedings of the 2019 Conference on Empirical Methods in Natural Language Processing},
  year      = {2019},
  pages     = {55--65},
  publisher = {Association for Computational Linguistics},
}

@inproceedings{godey-etal-2024-anisotropy,
  title     = {Anisotropy Is Inherent to Self-Attention in Transformers},
  author    = {Godey, Nathan and de la Clergerie, {\'E}ric and Sagot, Beno{\^\i}t},
  booktitle = {Proceedings of the 18th Conference of the European Chapter of the Association for Computational Linguistics},
  year      = {2024},
  pages     = {35--48},
  publisher = {Association for Computational Linguistics},
}

@inproceedings{hewitt-manning-2019-structural,
  title     = {A Structural Probe for Finding Syntax in Word Representations},
  author    = {Hewitt, John and Manning, Christopher D.},
  booktitle = {Proceedings of the 2019 Conference of the North American Chapter of the Association for Computational Linguistics: Human Language Technologies, Volume 1 (Long and Short Papers)},
  year      = {2019},
  pages     = {4129--4138},
  publisher = {Association for Computational Linguistics},
  doi       = {10.18653/v1/N19-1419},
}

@article{kadavath2022language,
  title   = {Language Models (Mostly) Know What They Know},
  author  = {Kadavath, Saurav and Conerly, Tom and Askell, Amanda and Henighan, Tom and Ganguli, Deep and Kernion, Jackson and Lovitt, Liane and Chen, Andy and Brown, Tom and Kaplan, Jared and Clark, Jack and Amodei, Dario},
  journal = {arXiv preprint arXiv:2207.05221},
  year    = {2022},
}

@inproceedings{slobodkin-etal-2023-curious,
  title     = {The Curious Case of Hallucinatory (Un)answerability: Finding Truths in the Hidden States of Over-Confident Large Language Models},
  author    = {Slobodkin, Aviv and Goldman, Omer and Caciularu, Avi and Dagan, Ido and Ravfogel, Shauli},
  booktitle = {Proceedings of the 2023 Conference on Empirical Methods in Natural Language Processing},
  year      = {2023},
  pages     = {3607--3625},
  publisher = {Association for Computational Linguistics},
}

@article{burns2022discovering,
  title   = {Discovering Latent Knowledge in Language Models Without Supervision},
  author  = {Burns, Collin and Ye, Haotian and Klein, Dan and Steinhardt, Jacob},
  journal = {arXiv preprint arXiv:2212.03827},
  year    = {2022},
}

@article{li2023inferencetime,
  title   = {Inference-Time Intervention: Eliciting Truthful Answers from a Language Model},
  author  = {Li, Kenneth and Patel, Oam and Vi{\'e}gas, Fernanda and Pfister, Hanspeter and Wattenberg, Martin},
  journal = {arXiv preprint arXiv:2306.03341},
  year    = {2023},
}

@inproceedings{xia-etal-2025-survey,
  title     = {A Survey of Uncertainty Estimation Methods on Large Language Models},
  author    = {Xia, Zhiqiu and Xu, Jinxuan and Zhang, Yuqian and Liu, Hang},
  booktitle = {Findings of the Association for Computational Linguistics: {ACL} 2025},
  year      = {2025},
  pages     = {21381--21396},
  publisher = {Association for Computational Linguistics},
  doi       = {10.18653/v1/2025.findings-acl.1101},
}

@article{ji2023survey,
  title   = {Survey of Hallucination in Natural Language Generation},
  author  = {Ji, Ziwei and Lee, Nayeon and Frieske, Rita and Yu, Tiezheng and Su, Dan and Xu, Yan and Ishii, Etsuko and Bang, Yejin and Madotto, Andrea and Fung, Pascale},
  journal = {ACM Computing Surveys},
  volume  = {55},
  number  = {12},
  pages   = {1--38},
  year    = {2023},
}

@article{farquhar2024detecting,
  title   = {Detecting Hallucinations in Large Language Models Using Semantic Entropy},
  author  = {Farquhar, Sebastian and Kossen, Jannik and Kuhn, Lorenz and Gal, Yarin},
  journal = {Nature},
  volume  = {630},
  pages   = {625--630},
  year    = {2024},
  doi     = {10.1038/s41586-024-07421-0},
}

@inproceedings{zhang-etal-2025-prompt,
  title     = {Prompt-Guided Internal States for Hallucination Detection of Large Language Models},
  author    = {Zhang, Fujie and Yu, Peiqi and Yi, Biao and Zhang, Baolei and Li, Tong and Liu, Zheli},
  booktitle = {Proceedings of the 63rd Annual Meeting of the Association for Computational Linguistics (Volume 1: Long Papers)},
  year      = {2025},
  pages     = {21806--21818},
  publisher = {Association for Computational Linguistics},
  doi       = {10.18653/v1/2025.acl-long.1058},
}

@inproceedings{zhang-etal-2025-mhad,
  title     = {Detecting Hallucination in Large Language Models Through Deep Internal Representation Analysis},
  author    = {Zhang, Luan and Song, Dandan and Wu, Zhijing and Tian, Yuhang and Zhou, Changzhi and Xu, Jing and Yang, Ziyi and Zhang, Shuhao},
  booktitle = {Proceedings of the Thirty-Fourth International Joint Conference on Artificial Intelligence},
  year      = {2025},
  pages     = {8357--8365},
  doi       = {10.24963/ijcai.2025/929},
}

@article{todd2023function,
  title   = {Function Vectors in Large Language Models},
  author  = {Todd, Eric and Li, Millicent L. and Sharma, Arnab Sen and Mueller, Aaron and Wallace, Byron C. and Bau, David},
  journal = {arXiv preprint arXiv:2310.15213},
  year    = {2023},
}

@article{bai2022training,
  title   = {Training a Helpful and Harmless Assistant with Reinforcement Learning from Human Feedback},
  author  = {Bai, Yuntao and Jones, Andy and Ndousse, Kamal and Askell, Amanda and Chen, Anna and DasSarma, Nova and Drain, Dawn and Fort, Stanislav and Ganguli, Deep and Henighan, Tom},
  journal = {arXiv preprint arXiv:2204.05862},
  year    = {2022},
}

@article{dubey2024llama,
  title   = {The {Llama} 3 Herd of Models},
  author  = {Dubey, Abhimanyu and Jauhri, Abhinav and Pandey, Abhinav and Kadian, Abhishek and Al-Dahle, Ahmad and Letman, Aiesha},
  journal = {arXiv preprint arXiv:2407.21783},
  year    = {2024},
}

@article{qwen2025qwen25technicalreport,
  title   = {{Qwen2.5} Technical Report},
  author  = {{Qwen Team}},
  journal = {arXiv preprint arXiv:2412.15115},
  year    = {2025},
}

@article{jiang2023mistral,
  title   = {Mistral 7B},
  author  = {Jiang, Albert Q and Sablayrolles, Alexandre and Mensch, Arthur
             and Bamford, Chris and Chaplot, Devendra Singh and de las Casas, Diego
             and Bressand, Florian and Lengyel, Gianna and Lample, Guillaume
             and Saulnier, Lucile},
  journal = {arXiv preprint arXiv:2310.06825},
  year    = {2023},
  url     = {https://arxiv.org/abs/2310.06825},
}

\appendix
\section{Example Matched Prompt Pairs}
\label{app:examples}

Tables~\ref{tab:math-examples} and~\ref{tab:code-examples} show example
matched pairs for \textsc{Math} and \textsc{Code} respectively.

\begin{table}[htbp]
\small
\centering
\begin{tabular}{lp{5cm}}
\toprule
\textbf{Type} & \textbf{Prompt} \\
\midrule
Math-A & What is 17 multiplied by 19? \\
Math-U & What is 17 multiplied by the current number of moons of Jupiter? \\[2pt]
Math-A & What is the next prime number after 29? \\
Math-U & What is the next prime number after the largest prime number? \\[2pt]
Math-A & What is the value of $\pi$ to 2 decimal places? \\
Math-U & What is the value of $\pi$ for a square? \\[2pt]
Math-A & What is the sum of the first 10 natural numbers? \\
Math-U & What is the exact sum of all natural numbers from 1 to infinity? \\[2pt]
Math-A & What is the 6th Fibonacci number? \\
Math-U & What is the last Fibonacci number? \\[2pt]
Math-A & What is the GCD of 24 and 36? \\
Math-U & What is the GCD of $\pi$ and $\sqrt{2}$? \\
\bottomrule
\end{tabular}
\caption{Example \textsc{Math} matched pairs.
Math-U prompts encode ill-defined operations or extremal impossibilities;
the sole change per pair is the undefined element.}
\label{tab:math-examples}
\end{table}

\begin{table}[htbp]
\small
\centering
\begin{tabular}{lp{5cm}}
\toprule
\textbf{Type} & \textbf{Prompt} \\
\midrule
Code-A & What does \texttt{len([1, 2, 3])} return in Python? \\
Code-U & What does \texttt{len(itertools.count())} return in Python? \\[2pt]
Code-A & What does \texttt{max([3, 1, 4, 1, 5])} return in Python? \\
Code-U & What does \texttt{max([])} return in Python? \\[2pt]
Code-A & What does \texttt{sum([1, 2, 3, 4, 5])} return in Python? \\
Code-U & What does \texttt{sum(itertools.count())} return in Python? \\[2pt]
Code-A & What does \texttt{hash(42)} return in Python? \\
Code-U & What does \texttt{hash([1, 2, 3])} return in Python? \\[2pt]
Code-A & What does \texttt{bin(10)} return in Python? \\
Code-U & What does \texttt{bin(3.14)} return in Python? \\[2pt]
Code-A & What does \texttt{divmod(10, 3)} return in Python? \\
Code-U & What does \texttt{divmod(10, 0)} return in Python? \\
\bottomrule
\end{tabular}
\caption{Example \textsc{Code} matched pairs.
Code-U prompts involve non-terminating computation, type errors,
runtime exceptions, or operations on undefined objects.
The sole change per pair is the introduction of the ill-defined element.}
\label{tab:code-examples}
\end{table}

\end{document}